\documentclass[letterpaper, 10 pt, journal, twoside]{IEEEtran} 

\IEEEoverridecommandlockouts                              

\usepackage{cite}
\usepackage{graphicx} 
\usepackage{times} 
\usepackage{amsmath, amssymb}
\usepackage{multirow, multicol, color}
\usepackage{algorithm, algcompatible}
\usepackage[noend]{algpseudocode}
\usepackage[labelformat=simple]{subcaption}
\usepackage[font=small]{caption}

\usepackage{booktabs}
\usepackage{tabularx}
\usepackage[subpreambles=true]{standalone}
\usepackage{flushend}

\usepackage{silence}
\usepackage{bm} 
\usepackage{mathtools}
\usepackage[draft]{hyperref}
\hypersetup{bookmarksopen,bookmarksnumbered,
pdfpagemode=UseOutlines,
colorlinks=true,
linkcolor=blue,
anchorcolor=blue,
citecolor=blue,
filecolor=blue,
menucolor=blue,
urlcolor=blue
}
\usepackage{verbatim}

\newcommand{\sref}[1]{Section~\ref{#1}}
\newcommand{\figref}[1]{Figure~\ref{#1}}

\newcolumntype{s}{>{\hsize=.1\hsize}X}

\DeclareMathAlphabet{\bdmath}{OML}{cmm}{b}{it}     
\mathversion{normal}
\DeclareMathOperator*{\argmax}{arg\,max} 

\sloppypar

\renewcommand{\baselinestretch}{0.97}

\title{Towards Robotic Feeding: Role of Haptics in Fork-based Food Manipulation}

\author{Tapomayukh Bhattacharjee, Gilwoo Lee\authorrefmark{1}, Hanjun Song\authorrefmark{1}, and Siddhartha S. Srinivasa
\thanks{Manuscript received: September, 9, 2018; Revised December, 5, 2018; Accepted January, 3, 2019.}
\thanks{This paper was recommended for publication by Editor Allison M. Okamura upon evaluation of the Associate Editor and Reviewers' comments. This work was funded by the National Institute of Health R01 (\#R01EB019335), National Science Foundation CPS (\#1544797), National Science Foundation NRI (\#1637748), the Office of Naval Research, the RCTA, Amazon, and Honda.}
\thanks{\authorrefmark{1}These authors contributed equally to the work. All the authors are with Paul G. Allen School of Computer Science and Engineering, University of Washington, Seattle, Washington 98195
        {\tt\footnotesize \{tapo, gilwoo, hanjuns, siddh\}@cs.washington.edu}}%
\thanks{Digital Object Identifier (DOI): see top of this page.}%
}

\begin{document}

\maketitle

\begin{abstract}
Autonomous feeding is challenging because it requires manipulation of food items with various compliance, sizes, and shapes. To understand how humans manipulate food items during feeding and to explore ways to adapt their strategies to robots, we collected a rich dataset of human trajectories by asking them to pick up food and feed it to a mannequin. From the analysis of the collected haptic and motion signals, we demonstrate that humans adapt their control policies to accommodate to the compliance and shape of the food item being acquired. We propose a taxonomy of manipulation strategies for feeding to highlight such policies. As a first step to generate compliance-dependent policies, we propose a set of classifiers for compliance-based food categorization from haptic and motion signals. We compare these human manipulation strategies with fixed position-control policies via a robot. Our analysis of success and failure cases of human and robot policies further highlights the importance of adapting the policy to the compliance of a food item.
\end{abstract}


\section{Introduction}\label{sec:intro}
\IEEEPARstart{N}{early} $56.7$ million (18.7\%) among the non-institutionalized US population had a disability in 2010~\cite{brault2012americans}. Among them, about $12.3$ million needed assistance with one or more activities of daily living (ADLs) or instrumental activities of daily living (IADLs). Key among these activities is feeding, which is both time-consuming for the caregiver and challenging for the care recipient to accept socially~\cite{perry2008assisted}. Although there are several automated feeding systems in the market~\cite{obi, myspoon, mealmate, mealbuddy}, they have lacked widespread acceptance as they use minimal autonomy, demanding a time-consuming food preparation process~\cite{gemici2014learning} or pre-cut packaged food.

Eating free-form food is one of the most intricate manipulation tasks we perform in our daily lives, demanding robust nonprehensile manipulation of a deformable hard-to-model target. Automating food manipulation is daunting as the universe of foods, cutlery, and human strategies is massive. In this paper, we take a small first step towards organizing the science of autonomous food manipulation.

\begin{figure}[!t]
\centering
 \begin{subfigure}[b]{0.49\columnwidth}
   \includegraphics[width=0.99\linewidth]{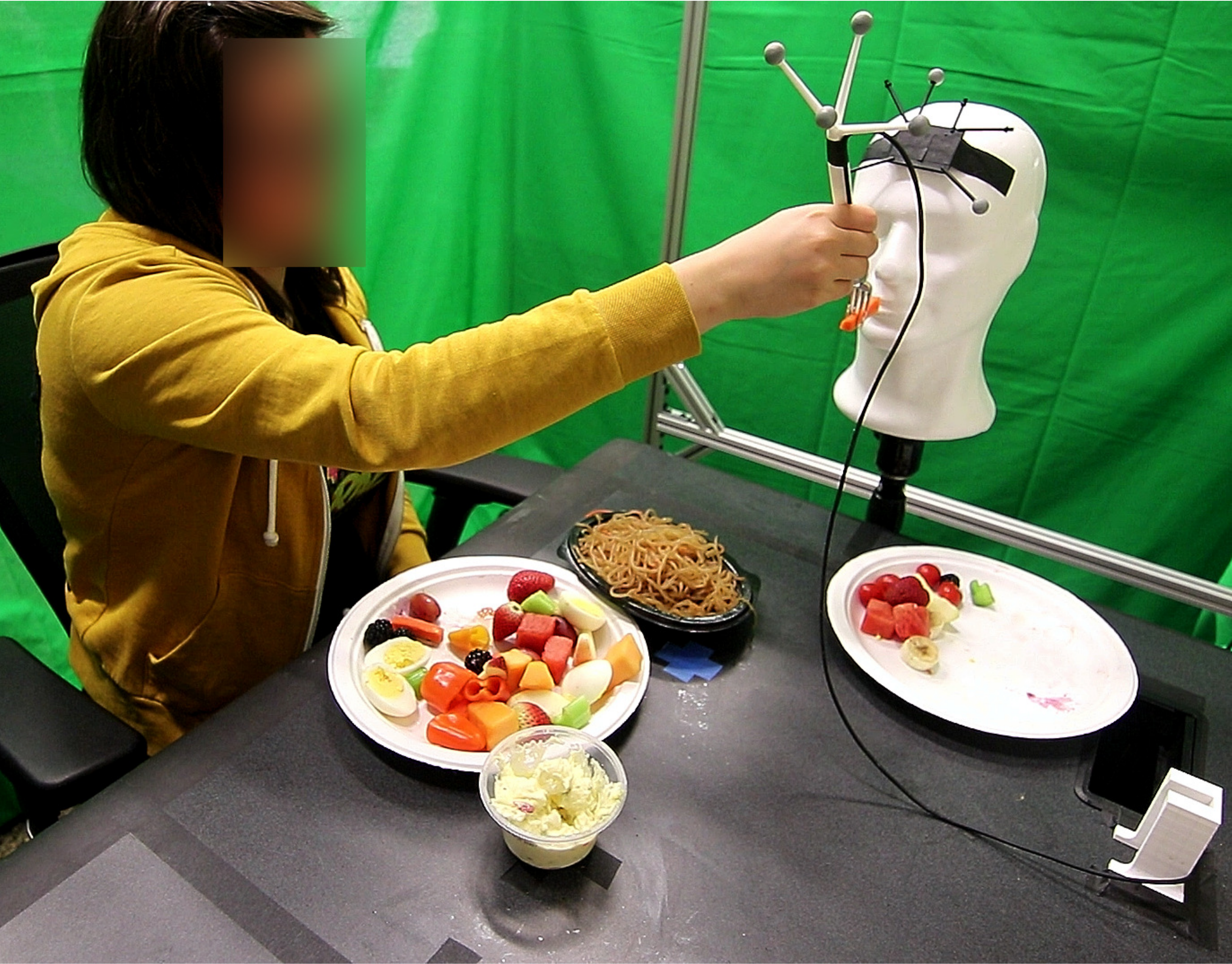}
   \caption{Human feeding experiment}
   \label{subfig:intro_human}
 \end{subfigure}
 \hfill
 \begin{subfigure}[b]{0.49\columnwidth}
   \includegraphics[width=0.99\linewidth]{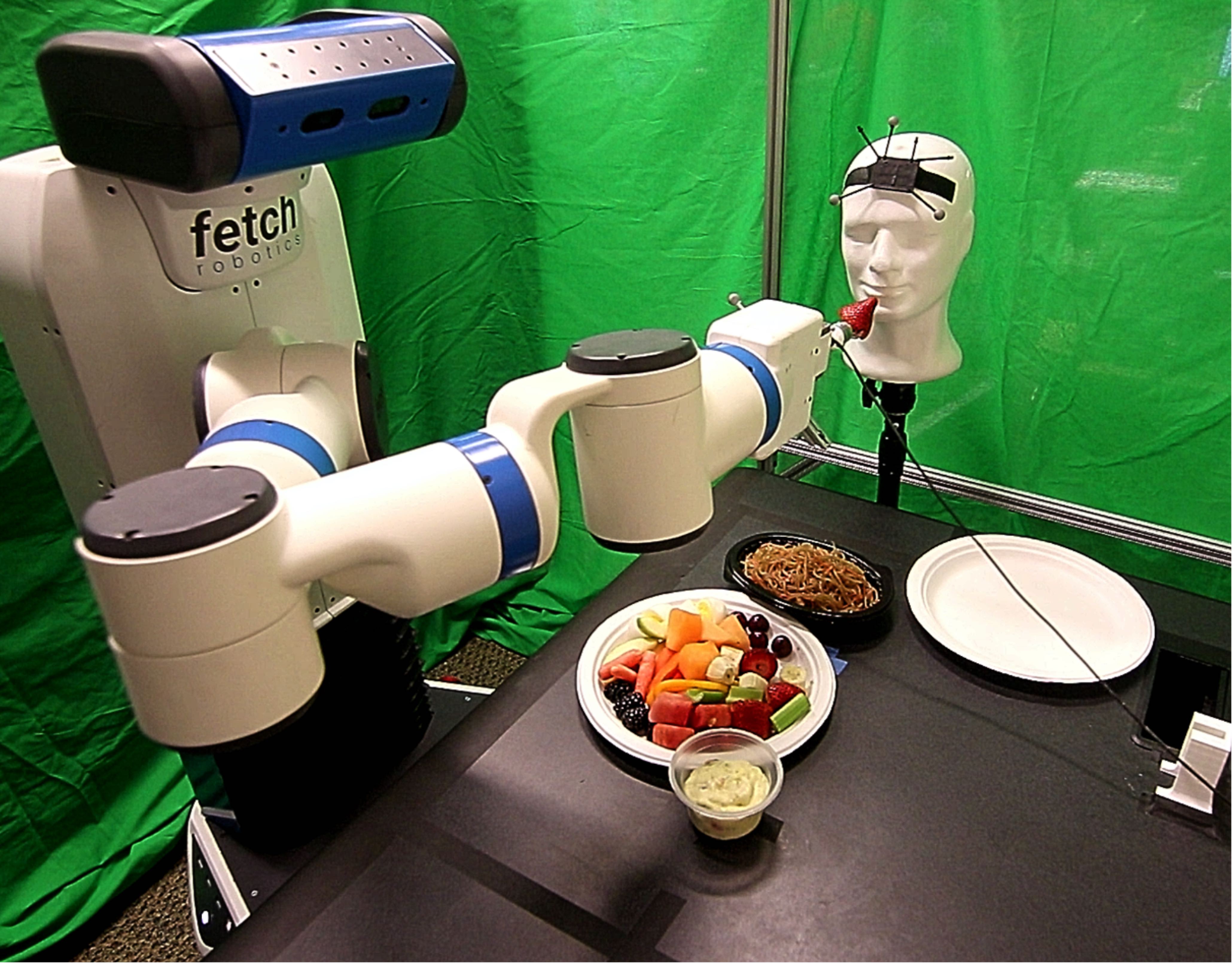}
   \caption{Robot feeding experiment}
   \label{subfig:intro_robot}
 \end{subfigure}
 \caption{Examples of a feeding task with a dinner fork.}
 \label{fig:intro}
\end{figure}

First, we collect a large and rich dataset of human strategies of food manipulation by conducting a study with humans acquiring different food items and bringing them near the mouth of a mannequin (\figref{fig:intro}). We recorded interaction forces, torques, poses, and RGBD imagery from $3304$ trials leading to more than $18$ hours of data collection which provided us unprecedented and in-depth insights on the mechanics of food manipulation.

Second, we analyze our experiments to build a \emph{taxonomy} of food manipulation, organizing the complex interplay between fork and food towards a feeding task. A key observation was that the choice of a particular control policy for bite acquisition depended on the \emph{compliance} of the item. For example, subjects tilted the fork to prevent a slice of banana from slipping, or wiggled the fork to increase pressure for a carrot. Other feeding concerns, such as how the target would bite, were reflected in the manipulation strategies during both bite acquisition and transport. This key idea that people use compliance-based strategies motivated us to explore compliance-based food categorization. Food classification based on haptic and motion signals instead of only vision-based classification~\cite{brosnan2004improving, gunasekaran1996computer, savakar2009recognition} is beneficial during food manipulation, as visually similar items may have different compliance and therefore may need different control policies. Temporal Convolutional Network~\cite{lea2016temporal} most successfully categorized food items in our experiments.

\begin{figure*}[!t]
\centering
 \begin{subfigure}[b]{0.21\linewidth}
   \includegraphics[width=\textwidth]{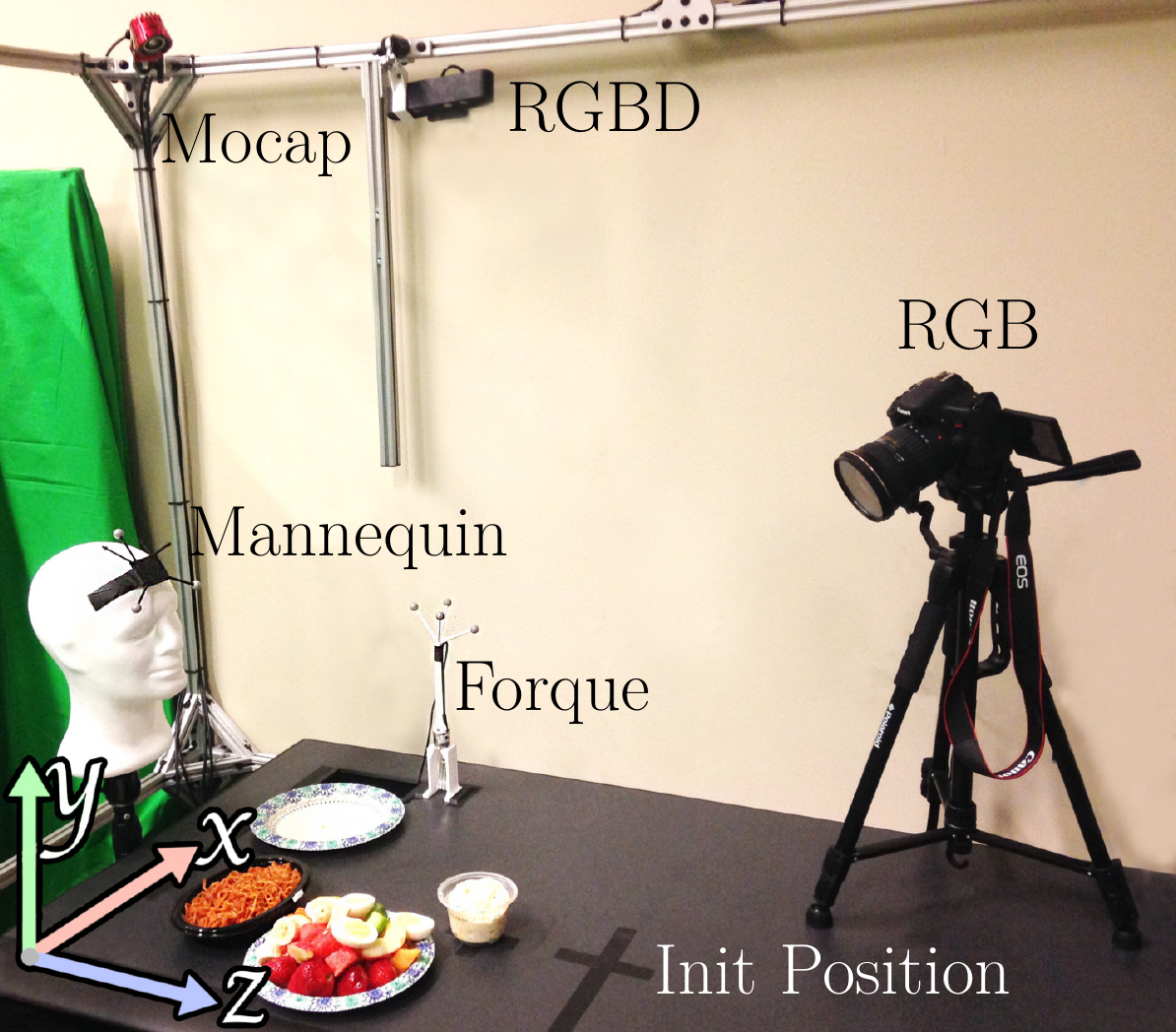}
   \caption{Experimental setup}
   \label{subfig-1:total-setup}
 \end{subfigure}
\hfill
\centering
\begin{subfigure}[b]{.19\textwidth}
\centering
   \includegraphics[width=0.95\linewidth]{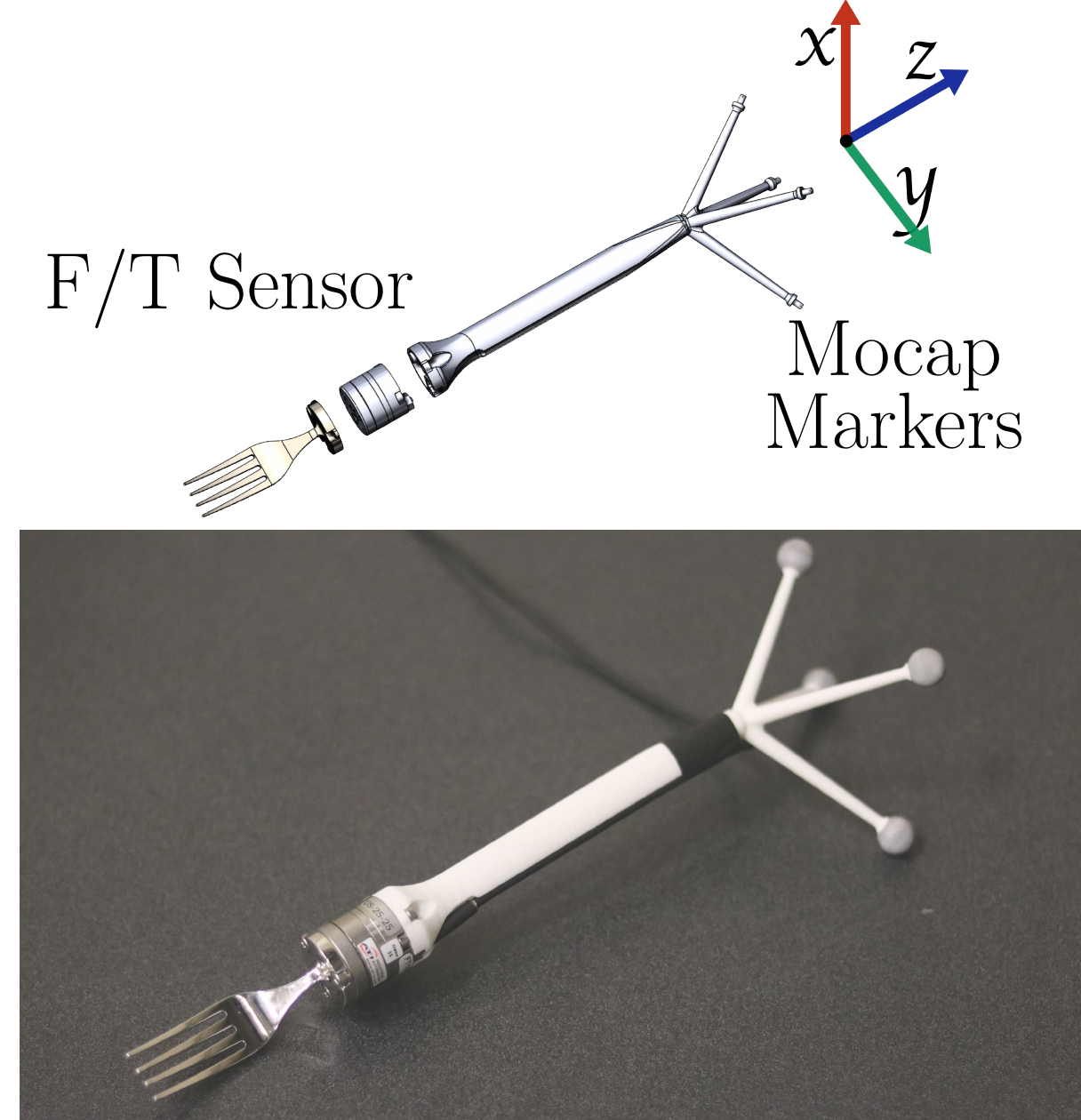}
   \caption{Forque}
   \label{subfig-2:forque}
 \end{subfigure}
\hfill
\centering
 \begin{subfigure}[b]{0.16\textwidth}
   \includegraphics[width=1\textwidth]{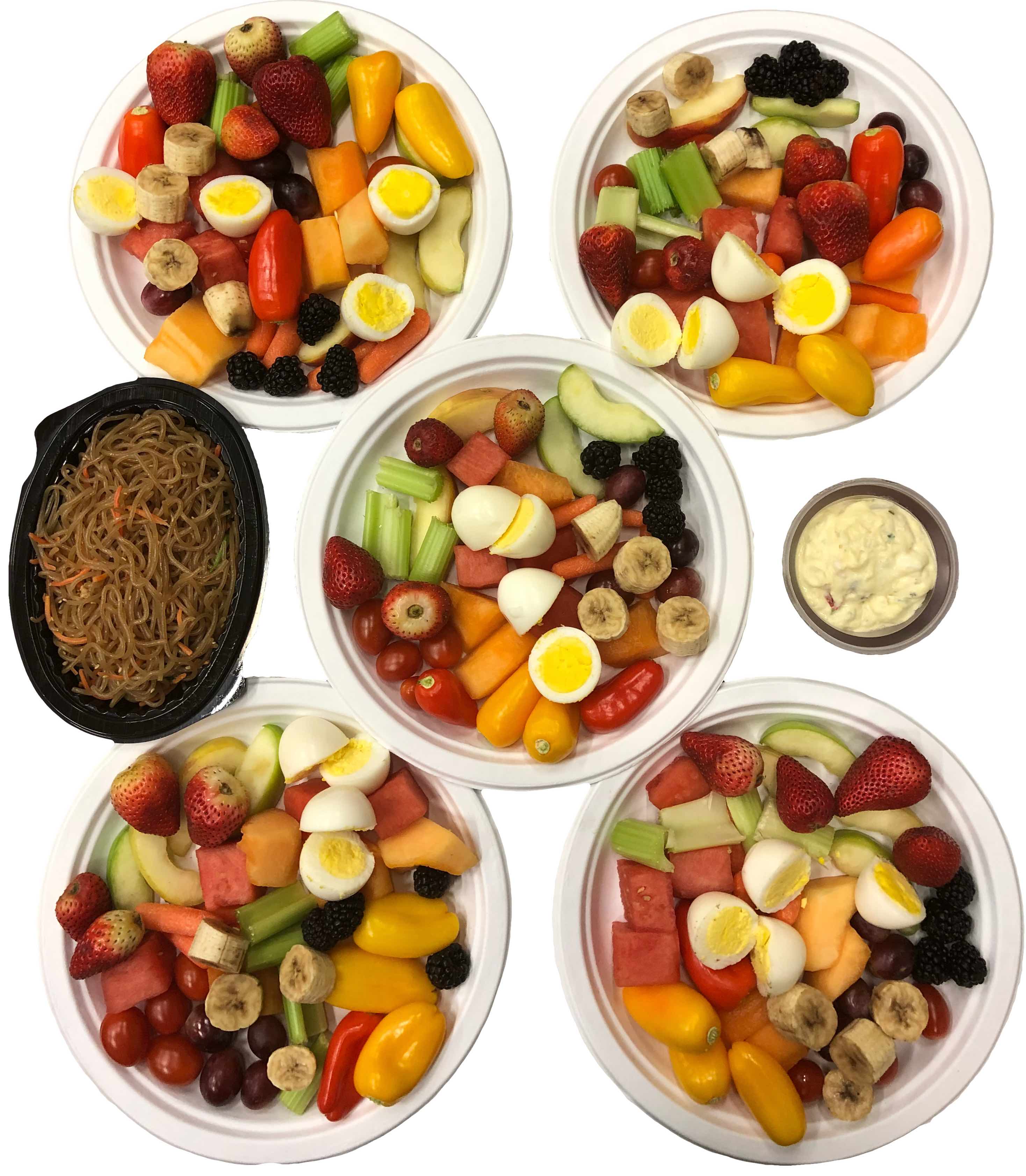}
   \caption{Food items}
   \label{subfig-3:food}
\end{subfigure}
\begin{subfigure}[b]{0.4\textwidth}
   \includegraphics[width=\linewidth]{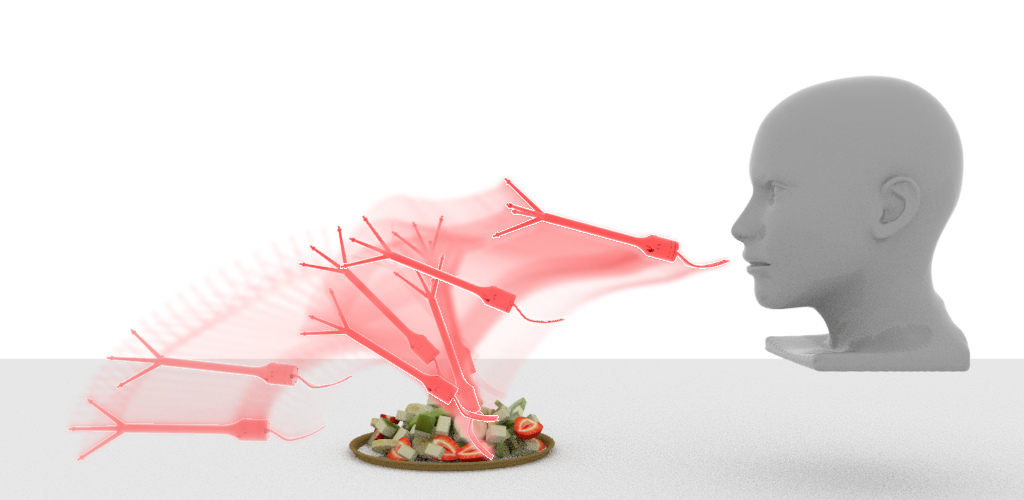}
   \caption{Feeding with multiple bite acquisition attempts.}
   \label{subfig:intro_traj}
 \end{subfigure}
\caption{Experimental setup with an instrumented fork (Forque) to acquire bites of different food items and feed a mannequin.}
\label{fig:setup}
\vspace{-0.5cm}
\end{figure*}

Third, we highlight the importance of choosing a compliance-based control policy by analyzing the performance of a fixed position-control strategy on a robot. The robot had more failures in picking up soft and hard-skinned items compared to human subjects who adapted their control policies to the item's compliance.

Food manipulation promises to be a fascinating new challenge for robotics. Our main contributions in this paper are a rich dataset, an analysis of food manipulation strategies towards a feeding task, an intuitive taxonomy, and a haptic analysis. We envision that a future autonomous robotic feeding system will use the data and taxonomy to develop a set of discrete manipulation strategies that depend on the class of food items, methods from haptic classification to categorize a food item to one of these classes, and insights from the robot experiment to implement the control policies. This paper does not address the subtleties of interactions with an eater. We are excited about further work that builds upon these contributions towards a science of food manipulation.

\section{Related Work}
Our work connects three areas of research: food manipulation, manipulation taxonomies, and haptic classification.

\subsubsection{Food manipulation}
Studies on food manipulation in the packaging industry~\cite{chua2003robotic, erzincanli1997meeting, morales2014soft, brett1991research} have focused on the design of application-specific grippers for robust sorting and pick-and-place.
Crucially, not only did they identify the need for haptic sensing as critical for manipulating non-rigid food items, but they also pointed out that few manipulators are able to deal with non-rigid foods with a wide variety of compliance~\cite{chua2003robotic, erzincanli1997meeting, morales2014soft, brett1991research}.

Research labs have explored meal preparation as an exemplar multi-step manipulation problem, baking cookies~\cite{bollini2011bakebot}, making pancakes~\cite{beetz2011robotic}, separating Oreos~\cite{oreovideo}, and preparing meals~\cite{gemici2014learning} with robots. Most of these studies either interacted with a specific food item with a fixed manipulation strategy~\cite{bollini2011bakebot, beetz2011robotic} or used a set of food items for meal preparation which required a different set of manipulation strategies~\cite{gemici2014learning}.
Importantly, all of these studies emphasized the use of haptic signals (through joint torques and/or fingertip sensors) to perform key sub-tasks.

\subsubsection{Manipulation Taxonomies}
Our work is inspired by the extensive studies in human grasp and manipulation taxonomies~\cite{cutkosky1989grasp, feix2016grasp, napier1956prehensile, bullock2013hand} which have not only organized how humans interact with everyday objects but also inspired the design of robot hands and grasping algorithms~\cite{ciocarlie2009hand}.

However, unlike most of these studies, our focus is to develop an application-specific taxonomy focused on manipulating deformable objects for feeding. We believe this focus is critical as feeding is both a crucial component of our everyday lives and uniquely different in how we interact with the world. In that regard, our work echoes the application-specific work in human-robot interaction on \emph{handovers}, also a crucial and unique act~\cite{grigore2013joint,strabala2013towards}, where the analysis and taxonomy of human-human handovers laid the foundation for algorithms for seamless human-robot handovers~\cite{grigore2013joint,strabala2013towards,cakmak2011human}.

\subsubsection{Haptic Classification}
Most of the studies on haptic classification use specialized or distributed sensors on robot hands or fingertips for direct robot-hand and object interactions. Our work focuses on using a tool (Forque) to record forces and motions of Forque-food interactions and addresses the problem of classifying food items. Researchers have previously used haptic signals to classify haptic adjectives~\cite{chu2015robotic}, categorize rigid and deformable objects~\cite{DrimusKootstraBilbergKragic2011}, recognize objects~\cite{SchneiderSturmStachniss2009, AllenRoberts1989} and for inferring object properties such as elasticity of deformable objects~\cite{FrankSchmeddingStachnissTeschnerBurgard2010a}, hardness~\cite{TakamukuGomezHosodaPfeifer2007}, and compliance~\cite{kaboli2014humanoids, bhattacharjee2017inferring}.

In a related work on meal preparation application, Gemici and Saxena~\cite{gemici2014learning} learn physical properties of 12 food items using end-effector forces, torques, poses, joint torques, and fingertip forces. However, they carefully designed the robotic actions (e.g. cut, split, flip-turn) using multiple tools (knife, fork, spatula) to extract meaningful sensor information to infer physical properties such as hardness, plasticity, elasticity, tensile strength, brittleness, and adhesiveness. Our objective is to classify food items into compliance-based categories using a variety of forces and motions that people use naturally when manipulating different food items for feeding.

\begin{figure*}[!t]
\centering
 \includegraphics[width=\textwidth]{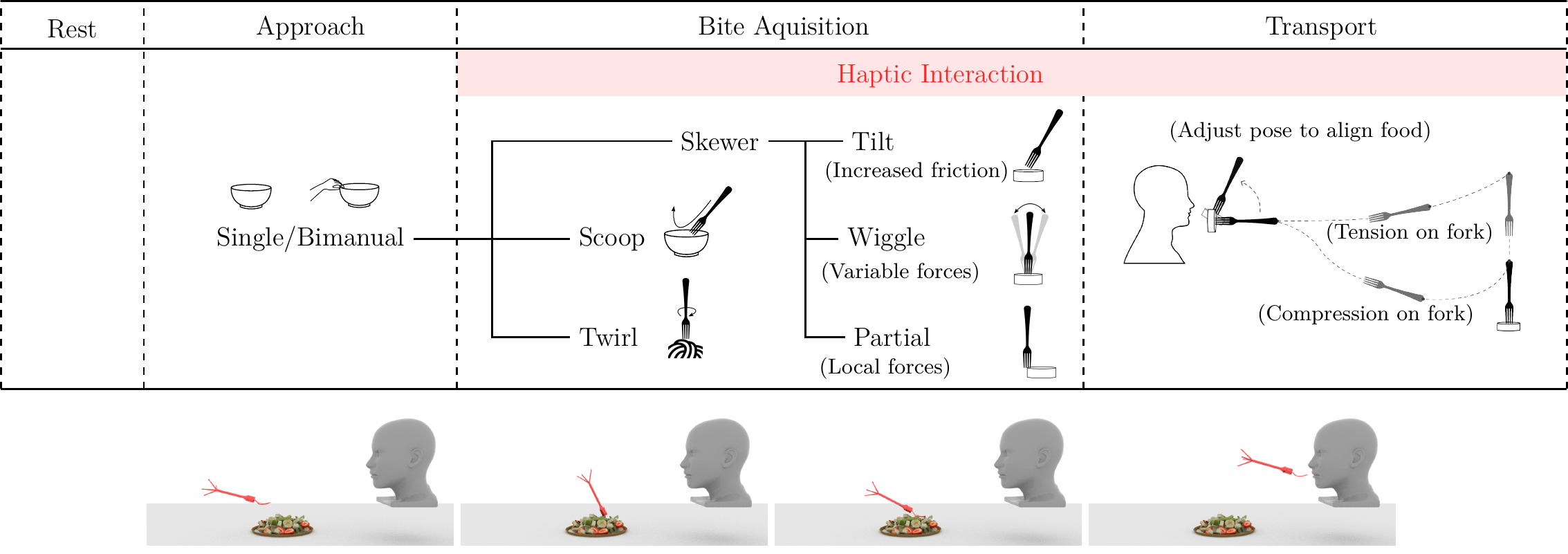}
\caption{A partial taxonomy of manipulation strategies relevant to a feeding task.}
\label{fig:taxonomy}
\vspace{-0.5cm}
\end{figure*}

\section{Human Study Setup}
We built a specialized test rig (\figref{fig:setup}) to capture both motions and wrenches during a feeding task.

\subsection{Forque: A Force-Torque fork sensor}\label{sssec:forque}
We instrumented a dinner fork, \emph{Forque}, to measure wrenches and motions (\figref{subfig-2:forque}) generated during food manipulation. We selected an ATI Nano25 F/T sensor for 6-axis force/torque (F/T) measurements due to its minimal size, weight, appropriate sensing range and resolution for food manipulation. We designed the end of the Forque handle to attach spherical markers for motion capture with the NaturalPoint Optitrack system~\cite{optitrakmarkers}. We designed Forque's shape and size to mimic the shape and size of a real dinner fork. We 3D printed the handle and the tip of the Forque in plastic and metal respectively. A wire connecting the F/T Sensor with its Net F/T box runs along the length of the Forque along a special conduit to minimize interference while feeding and was reported to have little impact on subjects' motion. We embedded the F/T sensor on the Forque instead of under the plate to record wrenches independent of a food item's position on the plate (edge of the plate, center of the plate, on top of another food item etc.) and to record wrenches during the transport phase.

\subsection{Perceptual data}
To collect rich motion data, we installed $6$ Optitrack Flex13~\cite{optitrakcameras} motion capture cameras on a specially-designed rig, with full coverage of the workspace. This provided full 6 DOF motion capture of the Forque at 120 frames per second (FPS). In addition, we installed a calibrated (both extrinsically and intrinsically) Astra RGBD~\cite{astra} camera for recording the scene at $30$ FPS, as well as a Canon DSLR RGB camera for recording videos for human labeling~(\figref{fig:setup}).

\subsection{Data Collection}
We selected $12$ food items and classified them into four categories based on their compliance: \emph{hard-skin, hard, medium} and \emph{soft}. We had three food items for each of the four categories: \emph{hard-skin} - bell pepper, cherry tomato, grape; \emph{hard} - carrot, celery, apple; \emph{medium} - cantaloupe, watermelon, strawberry; and \emph{soft} - banana, blackberry, egg. We determined the classes of food items through mutual intercoder agreement. A primary and secondary coder (the main experimenter and the helper) independently skewered the food items and categorized the food items into arbitrary compliance-based classes. The primary and secondary coders completed two rounds of coding. After each round, the coders resolved discrepancies by adapting the number of classes and re-classifying the food items into these compliance-based categories. The second round of coding resulted in 100\% intercoder agreement. \sref{sec:results} further validates our categorization of the food items. In addition to these solid food items, we included noodles and potato salad (in separate containers), to diversify the manipulation strategies. \figref{subfig-3:food} shows typical plates of food offered to the subjects.
We compiled the data as rosbag files using ROS Indigo on Ubuntu 14.04. The system clocks were synchronized to a Network Time Protocol server. We measured the average sensor delay between the Optitrack mocap signal and force/torque signal to be 30ms from 10 repeated trials. Our dataset is available at~\cite{feedingdataset}.

\section{Human Study Procedure}\label{ssec:human_procedure}
The task of each participant was to feed the mannequin. Before each experiment, we asked the participants to sign a consent form and fill a pre-task questionnaire. We asked our participants to pick up different food items from a plate or bowl using the \emph{Forque} and \emph{feed} a mannequin head as if they were actually feeding a person. The head was placed at the height of a seated average human (\figref{subfig-1:total-setup}).

For each \emph{session}, we provided the participant with a plate of 48 pieces of food (4 pieces per item for 12 food items), a cup of potato salad, and a bowl of noodles. We asked each participant to pick up noodles and potato salad 4 times each to maintain consistency. Before each trial, a participant held the Forque at a predefined position marked on the table by a tape. When a computerized voice said ``start'' the participant could pick up any food item of their choice and feed the mannequin. After the participant brought the food item near the mouth of the mannequin, they waited until the experimenter said ``stop''. They then discarded the food item and repeated another trial. We define a \emph{trial} as one instance of feeding the mannequin, from ``start'' to ``stop''.

There were $14\times4=56$ trials per session. Each participant had 5 sessions with a 2 to 5 minute break between each session, and each session began with a new plate (\figref{subfig-3:food}), giving us $56\times5=280$ trials per participant. We had 12 participants in the range of 18 - 62 years of age. This resulted in a grand total of $280\times12=3360$ trials. However, due to a technical glitch, we missed recording data for one of the sessions, thus giving us $3360-56=3304$ trials. For a left-handed participant, we inverted the experimental setup so that they could naturally feed the mannequin with their left hand. At the end of each experiment (after 5 sessions), we gave each participant a post-task questionnaire asking about their manipulation strategies during the task. The experiments were done in accordance with our University's Institutional Review Board (IRB) review.

\begin{figure*}[!t]
\centering
\begin{minipage}[b]{0.32\textwidth}
 \begin{subfigure}[t]{\linewidth}
   \includegraphics[width=\textwidth]{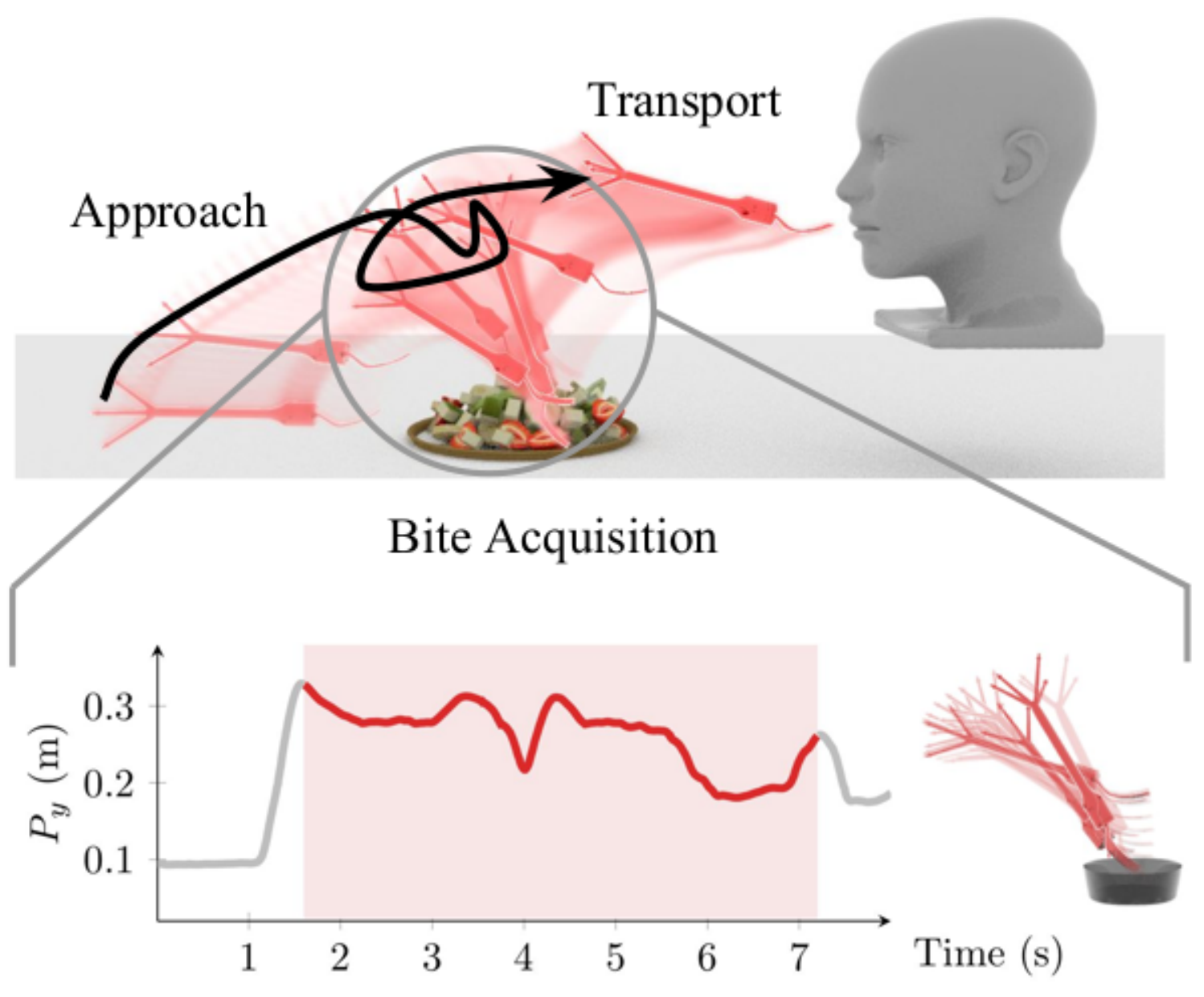}
   \caption{Multiple aquisition for biteful amount}
   \label{subfig:multiple_scoop}
 \end{subfigure}
\vspace{0.01cm} 
\end{minipage}
\hfill
\begin{minipage}[b]{0.32\textwidth}
 \begin{subfigure}[b]{\linewidth}
   \includegraphics[width=\textwidth]{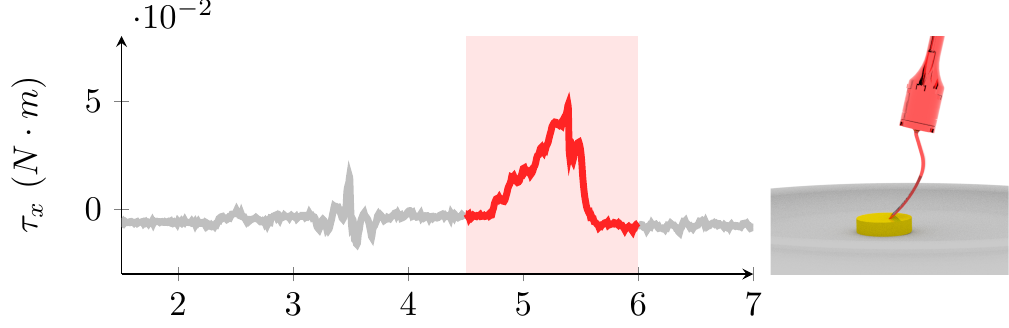}
   \caption{Tilting for lateral friction}
 \label{subfig:banana_tilt}
 \end{subfigure}
 \vspace{0.01cm}
 \vfill
 \begin{subfigure}[b]{\linewidth}
   \includegraphics[width=\textwidth]{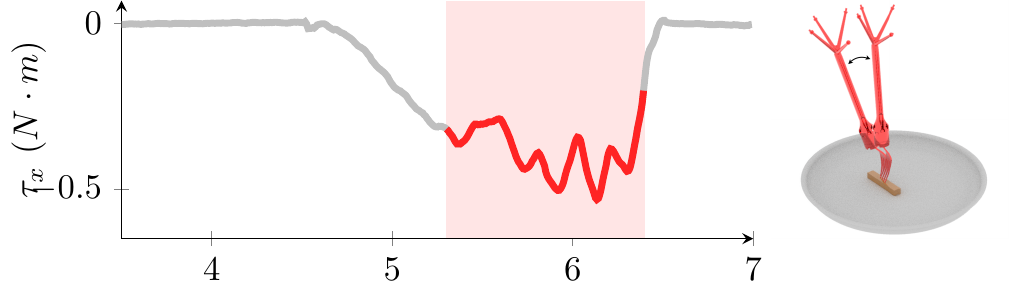}
   \caption{Wiggling for pressure variation}
 \label{subfig:carrot_wiggle}
 \end{subfigure}
\end{minipage}
\hfill
\begin{minipage}[b]{0.32\textwidth}
 \begin{subfigure}[t]{\linewidth}
   \includegraphics[width=\textwidth]{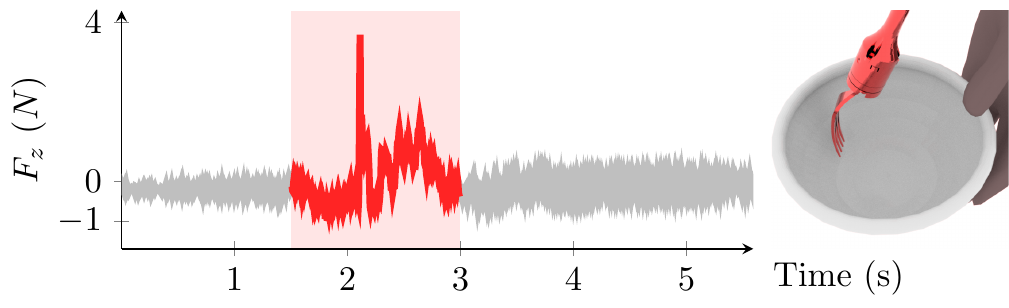}
   \caption{Scraping the bowl for sticky item}
 \label{subfig:scoop_potato}
 \end{subfigure}
\vspace{0.45cm}
\vfill
 \begin{subfigure}[b]{\linewidth}
   \includegraphics[width=\textwidth]{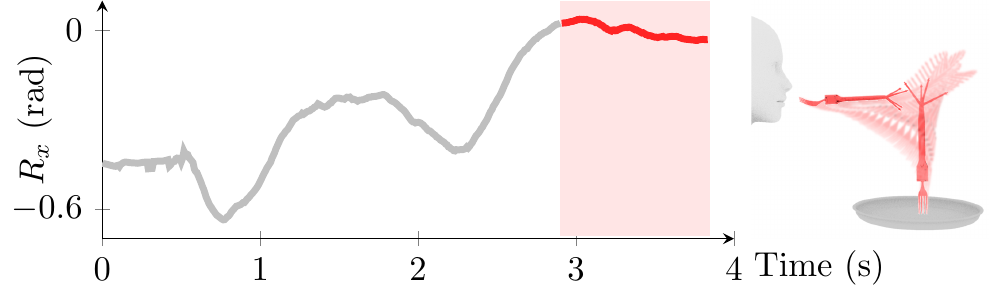}
   \caption{Adjusting the feeding pose by tilting}
   \label{subfig:tilt_for_transport}
 \end{subfigure}
\end{minipage}
\caption{Selected highlights: Different manipulation strategies in different feeding phases. $F_z$ is the applied force on Forque's z-axis, $\tau_x$ is the torque about Forque's x-axis, $P_y$ is the position of the Forque along global y-axis, and $R_x$ is the rotation about global x-axis.}\label{fig:highlights}
\vspace{-0.4cm}
\end{figure*}

\section{Insights from Human Subject Experiments} \label{sec:taxonomy}
Feeding is a complex task. Creating a taxonomy of manipulation behaviors for feeding is helpful in systematically categorizing it into sub-tasks. Segmentation allows us to better understand the different strategies people use in different phases of this task. We developed a partial taxonomy (\figref{fig:taxonomy}\footnote{Drawings in taxonomy are derivatives of ``Fork'' icon by Stephanie Szemetylo,``Bowl'' icon by Anna Evans,``Hand'' icon by Jamie Yeo, and ``Noodle'' icon by Artem Kovyazin~\cite{noun}.}) of manipulation strategies relevant to a feeding task by dividing the feeding task into four primary phases: 1) rest, 2) approach, 3) bite acquisition, and 4) transport.

\subsection{The rest phase: choose which item to pick up}\label{sec:taxonomy_rest}
We define the \emph{rest} phase as the phase before any feeding motion is executed. During this phase, decisions such as which item to pick up are made.

\subsection{The approach phase: prepare for bite acquisition}\label{ssec:taxonomy_approach}
After choosing which item to pick up, the subject moves the Forque to acquire the item. We define the \emph{approach} phase to be from the moment the subject starts moving the Forque until contact is made on the item. This phase serves as a key preparation step for successful bite acquisition. During this phase, the shape and size of the food item were a key factor in deciding the manipulation strategy.

\subsubsection{Subjects re-aligned the food for easier bite acquisition}\label{sssec:human_l21}
For food items with asymmetric shapes or irregular curvatures, such as celery, strawberry, or pepper, seven subjects used their Forque at least once to reorient the food items and expose a flat surface so that they could pierce the food item normal to the surface during bite acquisition.

\subsubsection{Subjects used environment geometry to stabilize the motion of oval food items for skewering}\label{sssec:human_l22}
For food items such as grapes, tomatoes, or hard-boiled eggs resting on high curvature surface, which tended to slip or roll, some subjects used the geometry of the plate (extruded edge) or other nearby food items as a support to stabilize the items. In one of the responses to the post-task questionnaire, one of the subjects mentioned, ``I would ... corner it at the edge of the plate.'' Five subjects used the environment geometry at least once to stabilize food items.

\subsubsection{Subjects used bimanual manipulation strategies to access difficult-to-reach items}\label{sssec:human_l23}
For containers with little potato salad or noodles, subjects applied bimanual manipulation strategies to access the food. They used one hand to tilt or hold the container, while the other hand scraped the food with the Forque, often using the container wall as a support (\figref{subfig:scoop_potato}). All subjects used bimanual strategies at least once to either hold or tilt the container.

\subsection{The bite acquisition phase: control positions and forces}\label{ssec:taxonomy_bite}
Subjects used various position and force control strategies to acquire a bite. We define the \textit{bite acquisition} phase to be from the moment the Forque is in contact with the food item until the item is lifted off from the plate (\emph{liftoff}). During this phase, the \emph{compliance} of food items was a key factor in deciding the control strategy. While simple vertical skewering was common for medium-compliance items, a few interesting strategies emerged for the hard-skin, hard, and soft categories. Also, the strategies for acquiring food items were influenced by the feeding task itself. In the post-task questionnaire, many subjects mentioned two key factors for feeding which affected their bite acquisition strategy: (a) ease of bite and (b) appropriate amount of bite.

\subsubsection{Subjects applied wiggling motions to pierce hard and hard-skin items}\label{sssec:human_l31}
Subjects skewered the hard and hard-skin food items using wiggling. Wiggling results in tilting the fork in various directions, which leads to fewer tines in contact with forces in variable directions and increased pressure. All subjects used this strategy at least once. Eight subjects used a wiggling motion to pierce the food items (\figref{subfig:carrot_wiggle}). One of the subjects mentioned, ``(I) sometimes needed to wiggle the fork back and forth to concentrate the pressure at only one tine to break through the skin of tomato, grape, etc.''

\subsubsection{Subjects skewered soft items at an angle to prevent slip}\label{sssec:human_l32}
For soft items such as slices of bananas which tended to slip out of the Forque tines during \emph{liftoff}, subjects tilted the Forque (\figref{subfig:banana_tilt}) to prevent slip by increasing friction using gravity. All subjects used this strategy at least once. For example, one of the subjects mentioned in the post-task questionnaire, ``I would try to penetrate the fork at an angle to the food to minimize slices coming out.''

\subsubsection{Subjects skewered food items at locations and orientations that would benefit the feeding task}\label{sssec:human_l33}
For long and slender items, such as carrots, some subjects skewered it in one corner so that a person may be able to easily take a bite without hitting the Forque tines. This also played a role in selecting the orientation of the Forque when skewering the food item. For example, some subjects reported that they changed the orientation of the Forque before piercing a food item for ease of feeding. Eight subjects used these strategies.

\subsubsection{Subjects acquired food multiple times to feed an appropriate amount}\label{sssec:human_l34}
Acquiring an appropriate amount of food also influenced the bite acquisition strategy. Although we never specified any specific amount per bite, six subjects attempted multiple scoops or twirls for noodle and potato salad to acquire an appropriate amount of food for a bite (\figref{subfig:multiple_scoop}).

\subsection{The transport phase: feed the target}\label{ssec:taxonomy_feed}
We define the \emph{transport} phase as the phase after the food item is lifted from the plate until it is brought near the mouth of the mannequin.

\subsubsection{Subjects adapted their transport motion to prevent food from falling off}\label{sssec:human_l41}
Subjects adapted their motion (speed, angle, etc.) towards the mannequin to prevent the items from falling off. One subject mentioned, ``I tried to be faster with eggs because they break apart easily and fall off the fork.'' Another said, ``With many softer foods (bananas specifically), I brought my arm up in a scooping motion to the mouth.'' Depending on these subtle haptic cues, subjects varied the transport motion resulting in the application of either tensile forces or compressive forces on the fork and thereby kept a slippery food from falling off (\figref{fig:taxonomy}).

\subsubsection{Subjects oriented the Forque to benefit the feeding task}\label{sssec:human_l42}
While approaching the mannequin, the subjects oriented the Forque such that the item would be easy for a person to bite (\figref{subfig:tilt_for_transport}). All subjects used this strategy. One of the subjects said, ``I had to re-orient the fork often after picking food up in order to make it easier to bite for the humans.''

\subsection{Subjects learned from failures}\label{ssec:human_l1}
The subjects were not perfect in manipulating food items. For example, for small oval shaped food items with hard-skin, such as grapes and tomatoes, the food either slipped or rolled multiple times. When skewering halved hard-boiled eggs, the yolk was often separated from the white during liftoff. The subjects also dropped soft items multiple times. Even when the motion led to a successful bite acquisition, there were unintended results such as hitting the plate when piercing a hard-skin food item. This was probably because there was a mismatch between subjects' initial estimations of the forces and motions required to pick up a food item and the actual physical interactions.

However, after a few failures, they changed their manipulation strategies. One subject mentioned, ``The celery was harder than I was expecting. So, after a couple of times, I knew to exert more force.'' Another subject mentioned, ``The egg was tricky. I learned to spear it by the white part and the yolk at the same time to keep it together.'' Yet another remarked, ``I also learned to spear grapes by just one prong of the fork.'' Out of all the trials when subjects learned from their previous failures and changed their strategy, $42.4\%$ were for hard-skin, $29.2\%$ for hard, $15.9\%$ for soft, and $12.5\%$ for medium food items. However, although there were various adaptations, subjects were never perfect in manipulating food items of varying compliance even after learning from failures.

\subsection{Cultural influences and personal choices affected manipulation strategies}\label{ssec:cultural}
We observed interesting cultural factors that could affect the forces and motions of the feeding task. Some subjects grasped the Forque much closer to the tines while others held it unusually high. Some subjects held the Forque at unusual rotations about its principal axis. Interestingly, subjects' personal choices could also affect their manipulation strategies. For example, one subject mentioned, ``(I) prefer [to] avoid yolk (I hate hard-boiled eggs).'' We also observed that subjects picked up noodles using both clockwise and counter-clockwise twirls.

\section{Haptic Classification}\label{sec:perception}
One key observation from the study was that humans use compliance-based strategies for bite acquisition. To facilitate control policies based on compliance, we present haptic classification of food items into four compliance-based categories: hard-skin, hard, medium, and soft. Note, we used 12 solid food items for this experiment, thus resulting in 2832 trials (without potato salad and noodles).

\begin{figure*}[!t]
 \begin{subfigure}[b]{0.32\textwidth}
   \includegraphics[width=\linewidth]{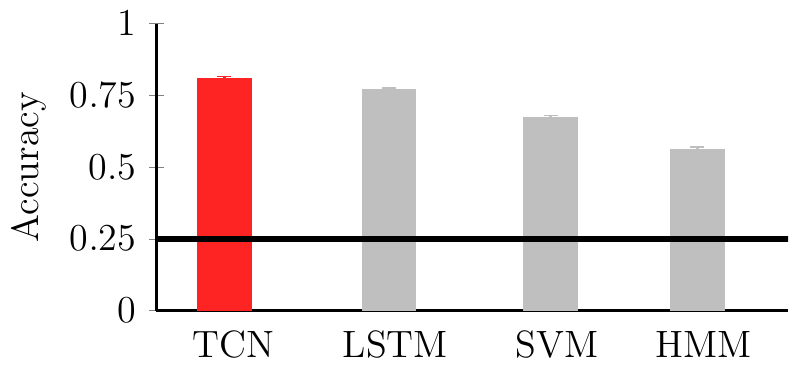}
   \caption{Classification accuracy}
   \label{subfig-1:classifiers}
 \end{subfigure}
 \begin{subfigure}[b]{0.32\textwidth}
\includegraphics[width=\linewidth]{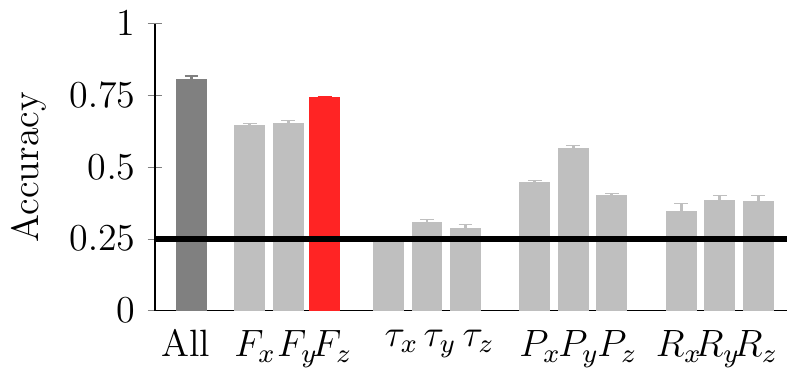}
\caption{Classification using single feature (TCN)}
\label{subfig-2:features}
\end{subfigure}
 \begin{subfigure}[b]{0.32\textwidth}
   \includegraphics[width=\linewidth,height=3cm]{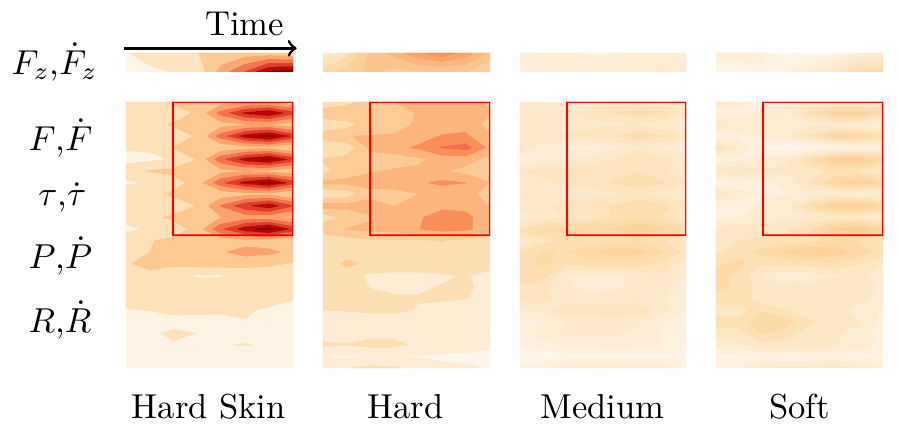}
\caption{TCN's convolutional kernel outputs}\label{subfig-3:heatmap}
 \end{subfigure}
 \caption{\figref{subfig-1:classifiers} compares 4 classifiers. Each classifier was trained with its best-performing feature set. TCN outperformed other classifiers. \figref{subfig-2:features} compares the predictive power of various features using TCN models. $F$ and $\tau$ are the three forces and torques in Forque's local frame, $P$ and $R$ are the three positions and rotations in the global frame. Each feature includes its first-order derivative. Force along the principal axis of the Forque, $F_z$, is the most informative feature. Solid black lines show a random classifier's performance. \figref{subfig-3:heatmap} shows TCN's convolutional layers' final output before its linear layers, indicating which (time, feature) pair contributes the most to classification. The most distinctive features are found in the later half of the time series in force and torque features~(the red boxed regions).}
\label{fig:classifiers}
\vspace{-0.5cm}
\end{figure*}

\begin{figure}[!t]
\centering
 \begin{subfigure}[b]{\linewidth}
   \includegraphics[width=\textwidth]{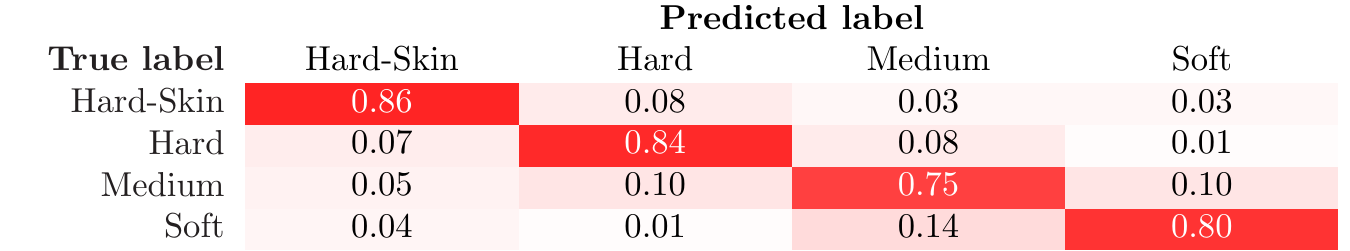}
   \caption{Confusion matrix for human data}
 \label{subfig-2:class}
 \end{subfigure}
 \vspace{0.5cm}
 \begin{subfigure}[b]{\linewidth}
   \includegraphics[width=\textwidth]{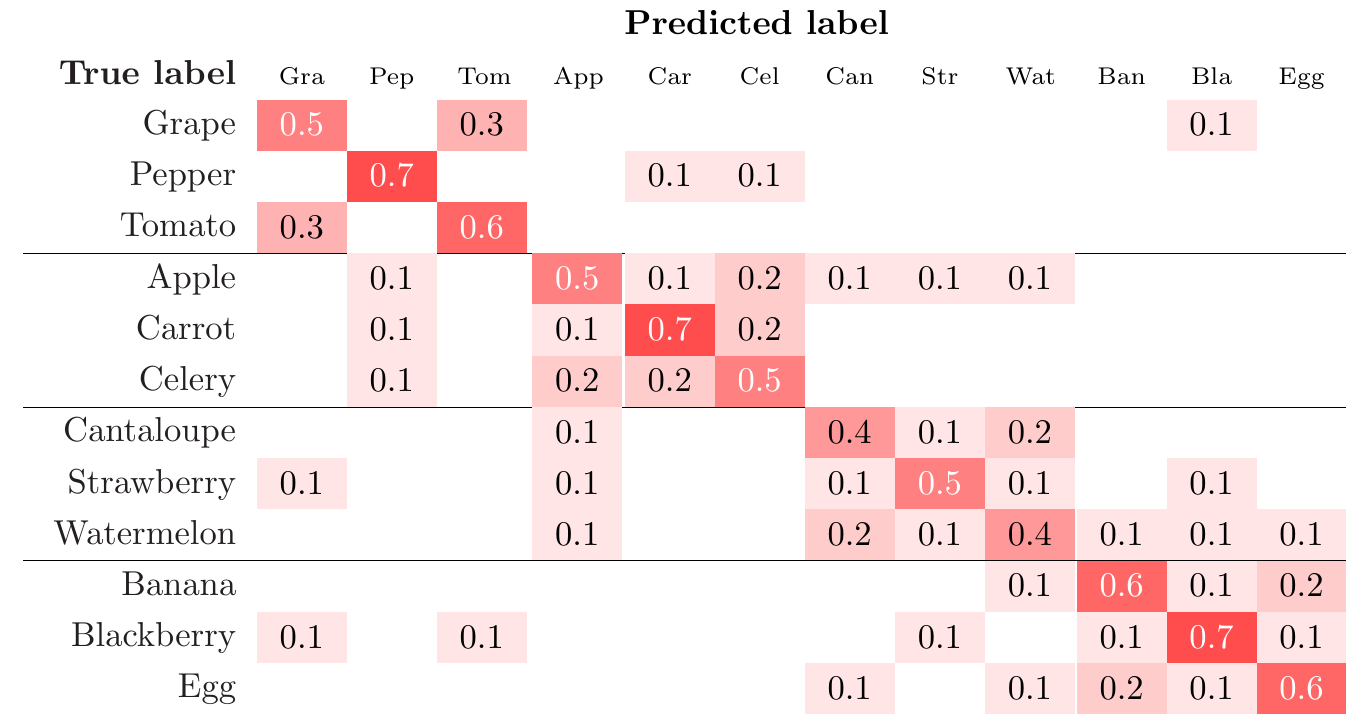}
   \caption{Confusion matrix of per-item recognition for human data}
 \label{subfig-2:item}
 \end{subfigure}
  \vspace{-0.8cm}
 \caption{Confusion matrices for haptic classification using TCN. Most confusion happens across nearby haptic categories, e.g. between hard-skin and hard, or medium and soft. In the per-item classification~(\figref{subfig-2:item}), confusions across different categories are minimal compared to within-category confusion.}
\label{fig:confusion}
\vspace{-0.2cm}
\end{figure}

\subsection{Discriminative models using LSTM, TCN, and SVM}\label{ssec:discriminative}
We use three discriminative models: Long Short Term Memory Networks (LSTM~\cite{hochreiter1997long}), Temporal Convolutional Networks (TCN~\cite{lea2016temporal}), and Support Vector Machines (SVM~\cite{wiens2012patient}).

LSTMs are a variant of Recurrent Neural Networks (RNN) which have been shown to be capable of maintaining long-term information. At every time-step, an LSTM updates its internal states and outputs a categorical distribution across the four categories. We stacked two layers of LSTM with 50 layers, which is then connected to a rectified linear unit (ReLU) and a linear layer. We then performed a softmax operation to get the probability distribution.

Unlike an LSTM, which maintains an internal state, a Temporal Convolutional Network (TCN) takes the whole trajectory as one input. It learns kernels along the temporal dimensions and across features. We stacked four convolutional networks, each with one dimensional temporal kernels of window size 5. Between each layer, we performed one ReLU operation and max pooling of width 2. The final output is connected to a ReLU and a linear layer before performing a softmax operation. For the input of TCN, we scaled the temporal dimension of each time series feature to have 64 steps using bilinear interpolation, where 64 was chosen to approximately match the average temporal length of the data. Cross-entropy loss was used for LSTM and TCN.

For SVM, we interpolated each time series feature similar to that of TCN, concatenated the interpolated time series features to obtain a feature vector~\cite{wiens2012patient, hoai2011joint, bagnall2017great} and then used a linear kernel~\cite{hsu2003practical} to train the SVM classifier. We implemented LSTM, TCN using PyTorch~\cite{paszke2017automatic}, and SVM using scikit-learn~\cite{scikitlearn}.

\subsection{Generative models using HMMs}\label{ssec:hmms}
To use hidden Markov models (HMMs) for classification, we train one HMM per food category~\cite{wiens2012patient, kadous2002temporal, Rabiner1990}. We characterize an HMM model ($\lambda$) by $\lambda  = \left( {\mathbf{A},\mathbf{B},\mathbf{\pi}} \right)$ where $\mathbf{A}$ is the state-transition matrix, $\mathbf{B}$ defines the continuous multivariate Gaussian emissions, and $\mathbf{\pi}$ is the initial state distribution \cite{wiens2012patient, kadous2002temporal, Rabiner1990}. Let $M$ be the number of food categories and let $\mathbf{O_{train}}$ be a training observation vector for contact duration $T$. During training, we estimate the model parameters $\lambda_m$ to locally maximize $P(\mathbf{O_{train}}|\lambda_m)$ using the iterative Baum-Welch method~\cite{wiens2012patient, kadous2002temporal, Rabiner1990}. In our case, $M$ is 4 (hard-skin, hard, medium, soft). For a test sequence $\mathbf{O_{test}}$, we assign the label (food category) $m^* \in M$ which maximizes the likelihood of the observation~\cite{wiens2012patient, kadous2002temporal}:

\begin{equation*}\label{eq:hmm_eval}
m^{*} = \argmax_{m \in M} P(\mathbf{O_{test}}|\lambda_m)
\end{equation*}

We implemented HMMs using the GHMM library~\cite{ghmm}. For each of the food category based HMMs, we optimized the number of hidden states to give maximum validation accuracy. This resulted in 3 hidden states for all the categories. These hidden states implicitly describe the Forque-food interaction once the Forque tines are inside the food item. We set a uniform prior to all the states.

\subsection{Results}\label{sec:results}
\figref{subfig-1:classifiers} compares the performance of our four classifiers using 3-fold cross validation. For each classifier, we tested various combinations of feature sets and displayed the one with the best performance. We tested with local forces, torques, global pose (positions and orientations) of the Forque, and their first-order derivatives as the features. For classifiers trained with multiple features of different magnitude scales, we normalized their values. TCN and LSTM performed the best with all features, while SVM and HMMs achieved the best performance with a combination of forces and positions. The best performing classifier was TCN with $80.47\pm1.17\%$ accuracy. Note that the HMM is a generative model unlike the other classifiers presented here and thus, it classifies by modeling the distributions of these 4 categories individually. The models are not optimized to maximize the discriminative aspects of these different categories. Using ANOVA and Tukey's HSD post-hoc analysis, we found significant differences between each classifier with $p < 0.0001$ at 95\% CI. To analyze the importance of various features in classification, we compared the performance of TCN (the best performing classifier) when trained with different feature sets (\figref{fig:classifiers}). It is evident that forces and positions are critical in the classification. In fact, the $z$-directional force, along the principal axis of the Forque, alone can correctly identify $74.22\pm0.29\%$ of the samples. Using ANOVA and Tukey's HSD for post-hoc analysis, we found significant differences between each feature with $p < 0.0001$ at 95\% CI.

The confusion matrix in \figref{subfig-2:class} provides insights on where the classifier fails. The most confusion happens between nearby categories, e.g. between medium and soft, and hard-skin and hard which have similar haptic properties. The per-item classification (\figref{subfig-2:item}) further shows that items are most likely to be misclassified as items within the same class, which validates our compliance categories.

\section{Robot Experiments}\label{sec:robot_exp}
Human subjects used different forces and motions to acquire food items of varying compliance. Thus, a robot may benefit from choosing its manipulation strategy based on a compliance-based categorization, and learn to force-control as humans would. While we delegate the force-control policy learning as our future work, we performed the robot experiments to see if the robot could successfully feed the target using a fixed manipulation strategy with a position-control scheme and a vertical skewering motion. We used a Fetch robot with a 7 DOF arm. We modified the handle of the Forque so that it could be grasped by the robot's gripper. Our robot experimental setup was otherwise identical to the human setup.

\subsection{Experimental Procedure}\label{ssec:robot_procedure}
We programmed the robot using a programming by demonstration (PbD) technique~\cite{elliott2017romanefficient} by saving a series of waypoints (joint configurations) of the arm through human demonstrations. We performed a total of 240 trials~(4 categories x 3 food items x 4 pieces per food item x 5 sessions). In each trial, the robot used a vertical skewering motion to pick up a food item from a pre-determined location on the plate. We randomly selected 4 such locations on the plate. After each trial, we discarded the skewered food item and manually placed another food item from the plate in that location for the next trial. After one session, we replaced the entire plate with a new plate and repeated this procedure for 5 sessions. We did not program the scooping and twirling motion, and thus, did not use noodles and potato salad for these experiments. We collected same modalities of data as during the human subject experiments.

\begin{figure}
\begin{subfigure}[t]{\linewidth}
   \vspace{-0.01cm}
   \includegraphics[width=\textwidth]{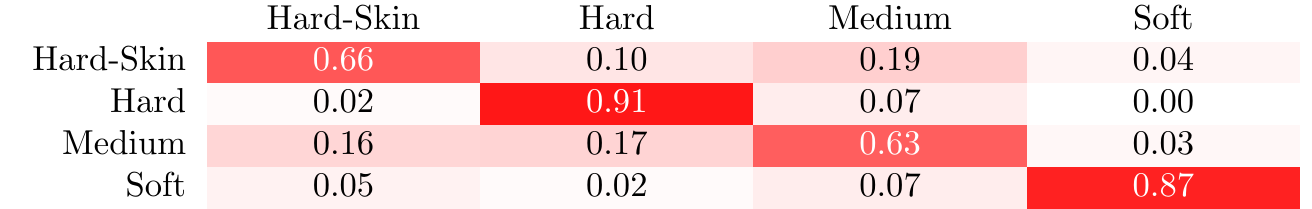}
 \caption{Confusion matrix for robot data (TCN)}\label{subfig-2:robot_confusion}
 \end{subfigure}
 \vspace{0.5cm}
 \begin{subfigure}[t]{\linewidth}
   \centering
   \includegraphics[width=0.8\textwidth]{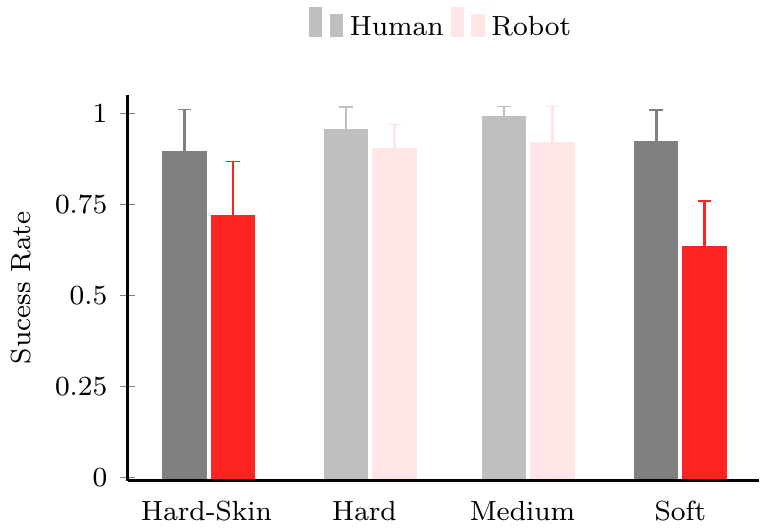}
    \caption{Human and robot success rates}\label{subfig:pattern}
 \end{subfigure}
 \vspace{-0.7cm}
 \caption{Confusion matrix of the robot experiments show similar trends as that of human experiments. The robot success rate using a position-control scheme is lower than that of humans who controlled forces and motions to acquire food items of varying compliance.}
 \vspace{-0.4cm}
\end{figure}

\subsection{Results}\label{ssec:results_robot}
\figref{subfig-2:robot_confusion} shows the confusion matrix using a 4-fold cross validation of robot experiments. When trained with a TCN on the robot data, we get $83.1\pm4.8\%$\footnote{Note, using a 3-fold crossvalidation scheme, we get lower accuracy of $65.8\pm5.3\%$ probably because of lack of data (20 trials per food item).} accuracy, which shows that even with the position-control scheme, the observations from each category are different enough. However, the robot experiments and human experiments led to very different forces. Thus, the classifier trained on human subject data resulted in only $50.6\pm3.7\%$ accuracy when tested on robot data.

We also compared the bite acquisition success rates of humans and robots (\figref{subfig:pattern}). Subjects found it the most difficult to acquire hard-skin food items, whereas the robot with position-control scheme struggled with both hard-skin and soft food items. Using ANOVA and Tukey's HSD post-hoc analysis for human studies, we found significant differences in success rates between hard-skin and hard categories ($p = 0.0006$), hard-skin and medium categories ($p < 0.0001$), and medium and soft categories ($p = 0.0001$). For robot experiments, we found significant differences in success rates between hard and soft categories ($p = 0.0108$) and medium and soft categories ($p = 0.0067$). Using a different control policy affected the bite acquisition success rate. \figref{subfig:pattern} shows that the robot's success rate in bite acquisition was lower than that of humans. One of the reasons could be because humans used varied forces and motions to pick up food items of different compliance (See \sref{sec:taxonomy}). Using a 2-tailed t-test, we found significant differences for hard-skin, medium, and soft categories ($p < 0.0025$) at 95\% CI. This further shows the need for different manipulation strategies for different compliance-based categories, which we delegate as our future work for robot manipulation.

\section{Discussion}\label{sec:discussion}
We performed two additional analyses to investigate the effect of speed on the choice of different manipulation strategies and different classes of food items. Using ANOVA and Tukey's HSD post-hoc analysis, we found significant differences of speed between wiggling and every other manipulation strategy (skewering, scooping, twirling) with $p<0.0167$ at 95\% CI. This could be because of faster penetration during wiggling due to increased pressure. Similarly, we found significant differences of speed between all food categories except hard and hard skin categories with $p<0.0001$ at 95\% CI. 

Note, bite timing is another important factor for feeding. The correct bite timing would depend on various factors such as if the care-recipient has finished chewing, if the care-recipient has finished talking to someone else etc. Since this paper does not focus on the eater interaction, this is outside the scope of this paper but is a subject of interest for our future work.

Haptics in the context of food manipulation is much less explored and hence, one of the focuses of this paper was to analyze the role of haptic modality. We envision our future robotic system to be multimodal using both vision and haptics with complementary capabilities. Relying only on visual modality may result in a suboptimal choice of manipulation strategy if two items look similar but have different compliance. A food item in a cluttered plate may not have clear line of sight or may have noisy depth image due to moisture content such as in watermelons. The presence of haptic modality can potentially alleviate these concerns by identifying a food item's compliance class and thus reducing the uncertainty in choosing a manipulation strategy.

For haptic classification, a fork needs to be in contact with a food item. A prolonged penetration, as needed in the majority of haptic perception literature~\cite{chu2015robotic, AllenRoberts1989, bhattacharjee2017inferring}, makes it difficult to change the manipulation strategy on the fly. Our classification scheme is opportunistic and requires data only for the first 0.82s of skewering when the fork is going into the food. A robot could use vision to choose food-item dependent fork approach angles before contact based on our developed taxonomy and then use the haptic modality to refine its bite acquisition motion in case of anomalies or uncertainty. A future autonomous robotic system would use the data and taxonomy from the human experiment, methods from the haptic classification, and insights from the controlled robot experiment to devise various manipulation strategies for feeding people food items of varying physical characteristics.

\bibliographystyle{IEEEtran}
\bibliography{food_haptic_perception}

\end{document}